# Physical Computing for Materials Acceleration Platforms

Erik Peterson
Alexander Lavin*

Pasteur Labs; Brooklyn, NY, USA
* Corresponding author: lavin@simulation.science

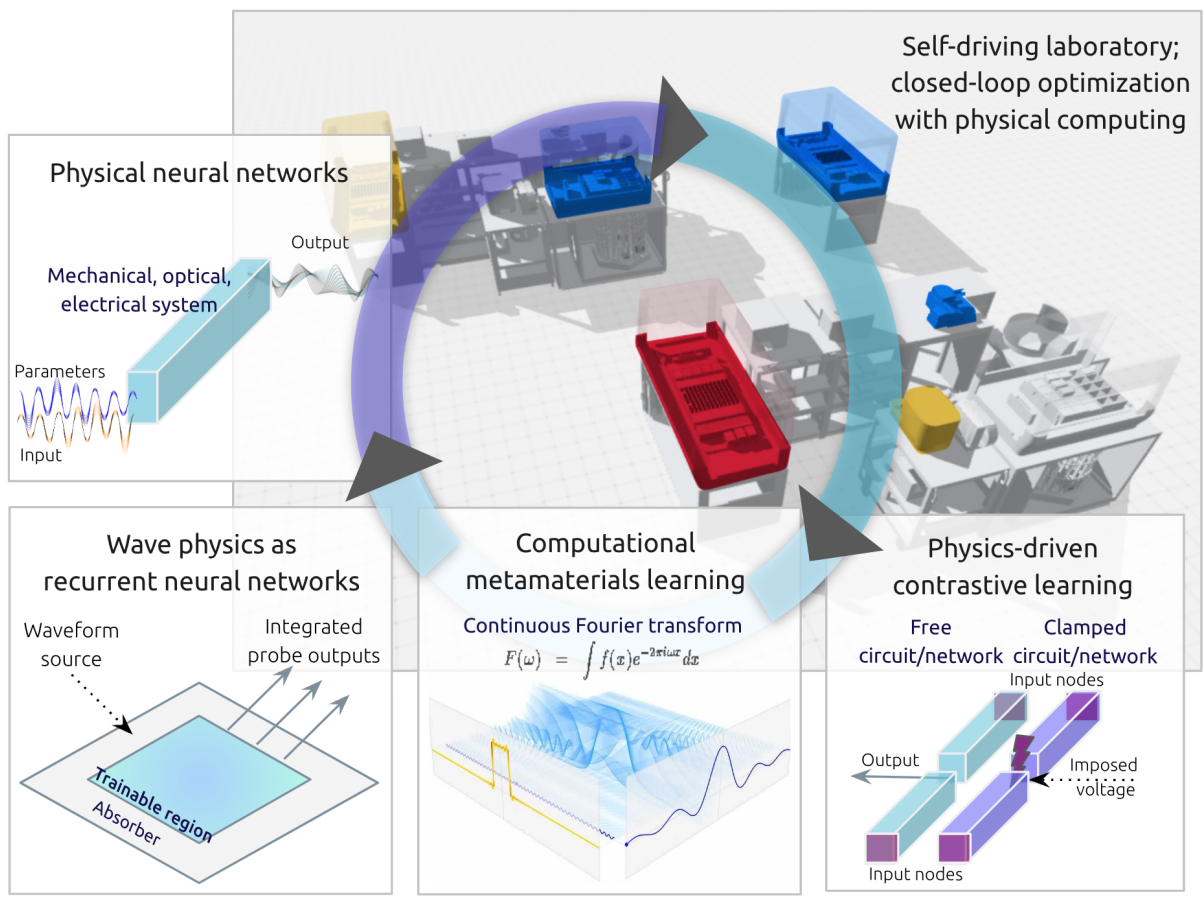

Nontraditional computing paradigm is prescribed for the forthcoming wave of materials acceleration platforms (MAP) and self-driving laboratories: physical computing (PC), to perform information processing and machine learning using naturally occurring physical processes for analog computation (rather than imperfect conversions to bits and data structures for digital or software computations). Four priority approaches are highlighted for PC-MAP, notably physical machine learning and computational metamaterials, that can outperform digital counterparts in self-driving lab workflows, and produce new dimensions of scientific methods driven by artificial intelligence.

**Abstract**

A "technology lottery" describes a research idea or technology succeeding over others because it is suited to the available software and hardware, not necessarily because it is superior to alternative directions — examples abound, from the synergies of deep learning and GPUs to the disconnect of urban design and autonomous vehicles. The nascent field of Self-Driving Laboratories (SDL), particularly those implemented as Materials Acceleration Platforms (MAPs), is at risk of an analogous pitfall: the next logical step for building MAPs is to take existing lab equipment and workflows and mix in some AI and automation. In this whitepaper, we argue that the same simulation and AI tools that will accelerate the search for new materials, as part of the MAPs research program, also make possible the design of fundamentally new computing mediums. We need not be constrained by existing biases in science, mechatronics, and general-purpose computing, but rather we can pursue new vectors of *engineering physics*[1] with advances in cyber-physical learning and closed-loop, self-optimizing systems. *Here we outline a simulation-based MAP program to design computers that use physics itself to solve optimization problems.* Such systems mitigate the hardware-software-substrate-user information losses present in every other class of MAPs and they perfect alignment between computing problems and computing mediums eliminating any technology lottery. We offer concrete steps toward early *Physical Computing (PC) -MAP* advances and the longer term cyber-physical R&D, which we expect to introduce a new era of innovative collaboration between materials researchers and computer scientists.

# Introduction

All modern scientific computing and machine learning (ML) schemes are based on general computing architectures like the modern CPU or GPU, including those essential to building MAP and other self-driving labs (SDLs). With multiple layers of abstraction between hardware, software, applications and users, computational schemes fundamentally suffer from a trade-off between accurately representing and measuring physical systems and overall computational and experimental complexity. This trade-off leads to practical information losses from *in situ* characteristics to *in silico* information. In theory if we had perfect measurement devices for any given property of a class of material — that is, perfect conversion from analog measurement (or signals) to digital information (bits), and vice-versa (which we denote A↔D)[1] — digital computing to approximate physical systems could be lossless. Yet in practice it is nearly impossible for any one A↔D conversion to be without error, and it is guaranteed impossible over a larger set of multiple materials let alone upwards of $10^{60}$ possible materials.[2] With the scaling of SDLs there are dozens to millions of measurements in a MAP pipeline, which not only need to be represented accurately but also filtered and classified for downstream use. The resulting series of conversion errors will cascade and accumulate to a nontrivial amount of information lost thereby skewing results. Current tools barely make practitioners aware of these errors,[3] let alone how to quantify or avoid them; the existing schemes cannot faithfully propagate errors for training,[4] uncertainties for probabilistic reasoning, and gradient-based information for learning across cyber-physical boundaries.

---

[1] ***Engineering physics***, or engineering science, refers to the study of the combined disciplines of physics, mathematics, chemistry, biology, and engineering, particularly computer, nuclear, electrical, electronic, aerospace, materials or mechanical engineering. By focusing on the scientific method as a rigorous basis, it seeks ways to apply, design, and develop new solutions in engineering. (source: https://en.wikipedia.org/wiki/Engineering_physics)

The ideal MAP information processing pipeline would be real-time, compact, parallelizable, and power efficient.[1] *Physical computing*, or information processing via physical mechanisms, may be able to offer these properties but has in the past been limited by the same reliability problems that limit analog computing more generally, such as non-trivial noise in physical elements.[5] Gradient-based information across hardware-software-substrate-human interfaces can be the key, made possible by physical computing. The use of differentiable computing along with distributed or connectionist representations[2] will allow the design of a new generation of reliable physical computers matched to the processing needs of self-driving laboratories.

*Assertion 1: Information loss across cyber-physical abstractions and boundaries is present in non-trivial amounts, and can introduce cascading and accumulating effects over the cyber-physical information processing pipelines.*

*Proposition 1: We look to design computers that use physics itself to propagate information from initial measurement to signal (pre-)processing to final classification or output.*

*Physical ML* describes a class of methods for exploiting the physical processes of laboratory mechanical and electromagnetic networks in tandem with various learning rules and data (simulated and real), where physical processes rather than abstract mathematical operations are trained. Physical ML systems have at least two significant advantages over conventional ML done on general purpose computers. First, by directly solving physical optimization problems using physical processes themselves, they can eliminate the need to generate abstract digital representations which cause information losses.[1] Second, they *require* we sidestep any technology lotteries (Fig. 1) by learning to align the computing material to the physical computing problem—from initial measurement to final output. Recently developed methods allow us to design practical, bespoke physical computers for machine learning. While the details of these methods vary (described below), they all employ data-driven connectionist representations akin to those seen in artificial neural networks. The representational degeneracy inherent in connectionist and distributed designs allows for physical analog computers that can overcome instabilities seen in classic analog computers,[5] and can parallelize large workloads because of implicit concurrency.[1]

One method of *Physical ML* is *Physics-driven contrastive learning*[7] which compares the response of a connectionist electronic circuit to two different boundary conditions and adjusts the degrees of freedom/weights; as demonstrated in Dillavou et al.[8] the physical imperative to minimize energy dissipation in the circuit carries out the forward calculation to "compute" the outputs within nanoseconds, while local rules that adjust the resistances of the edges take the place of backpropagation, obviating the need for a processor or memory storage because the mechatronics themselves encode memory. Another method is that of Hughes et al.[9] who identified a mapping between physical waves (propagating through any heterogeneous medium) and abstract recurrent neural networks (RNNs) and proved that the training mechanisms from neural networks can be applied to guiding wave propagation as physical computations. Their Wave-RNN passively processed signals and information in a native physical domain without the expense and errors associated with analog-to-digital conversion. As the authors note, "using physics to perform computation may inspire a new platform for analog machine learning devices, with the potential to perform computation far more naturally and efficiently than their digital counterparts."

---

[2] "The term **connectionism** is usually applied to neural networks. There are, however, many other models that are mathematically similar, including classifier systems, immune networks, autocatalytic chemical reaction networks, and others. In view of this similarity, it is appropriate to broaden the term connectionism. [One may] define a connectionist model as a dynamical system with two properties: (1) The interactions between the variables at any given time are explicitly constrained to a finite list of connections. (2) The connections are fluid, in that their strength and/or pattern of connectivity can change with time."[6]

A complementary approach is *physics-aware training* where the backpropagation algorithm is implemented *in situ* to train the hardware's physical transformations directly for performing desired computations.[4] The focus in the original work was on training existing hardware devices—i.e., simple mechanical, optical, and electrical systems. But by erasing the traditional software–hardware division, as we argue here, a new breed of physical computers and *physical neural networks (PNNs)* will let us opportunistically construct neural network hardware from virtually any controllable physical system(s) or material(s); controllable in this sense means the system can be parameterized to alter the physical transformations made on input data. PNNs can provide several of the benefits we highlight for MAPs: automatic mitigation of imperfections and noise because you're optimizing in the space of mechatronics, end-to-end propagation of errors (and potentially uncertainties), fast and energy efficient ML compared to conventional electronic processors, and the ability to endow physical systems with automatically designed physical functionalities. There exist many examples of MAPs producing physical structures for computing, namely in materials,[11–13] robotics,[14–16] and smart sensors.[17–19] Eliminating the need for analog-digital (A↔D) transforms means physical NNs and broader physical ML can have profound influence on new sensors and signal processors, such as learned sensing and closed-loop sensor design, both described below.

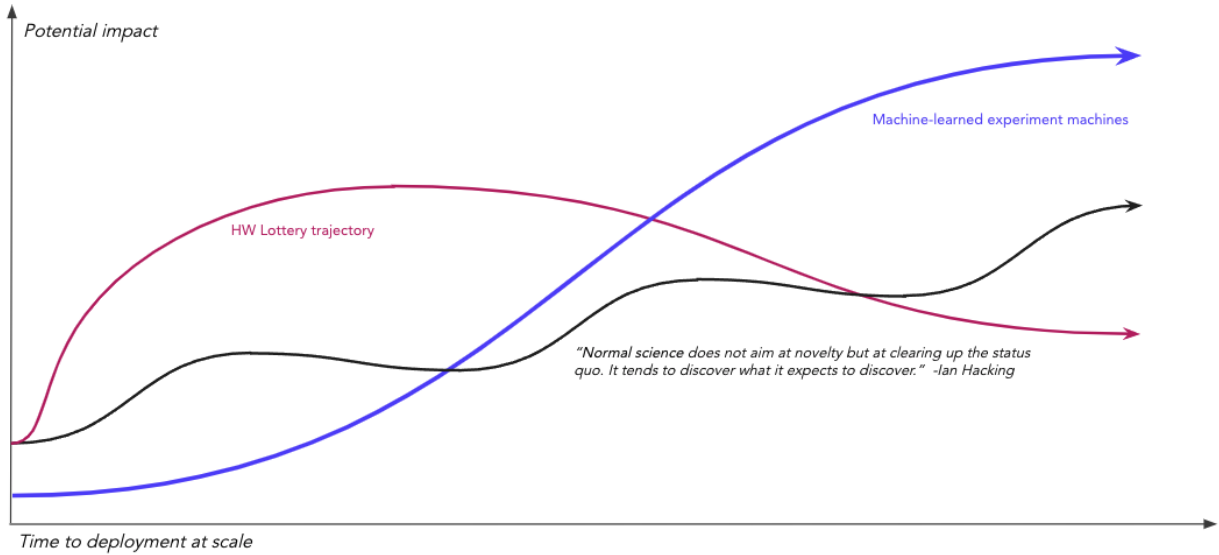

**Figure 1, Trajectories for the MAP field** — Visualization of impact over time of several MAP paths. Scientific progress occurs when there is a confluence of factors which allows the scientist to overcome the "stickyness" of the existing paradigm. The black line reflects *normal science for MAPs* (e.g. simply shoving AI into legacy machine workflows) which struggles to make non-trivial progress. The deep learning hardware lottery we alluded to suggests the speed at which paradigm shifts have happened in AI research have been disproportionately determined by the degree of alignment between hardware, software and algorithm, and thus any attempt to avoid hardware lotteries must be concerned with making it cheaper and less time consuming to explore different hardware-software-algorithm combinations.[10] A hypothetical *HW lottery for MAPs* trajectory is drawn in red. The bigger picture for AI-driven science can be realized with the PC-MAP methods we propose, labeled here as *machine-learned experiment machines* in purple.

Some computing functions, notably differentiation and integration tasks that are simple but nonetheless cost-inefficient in A↔D components, have been be implemented directly in the substrate. Such *computational metamaterials* implement single computing functionalities via light propagation in suitably engineered artificial photonic materials.[20] This class of wave-based "classical' analog computing can operate at the speed of light with a footprint only the size of a wavelength. And, not beholden to the fundamental thermal limits of Moore's law, passive-wave-based systems are massively parallelizable, and thus can provide magnitudes better efficiency in speed or power relative to conventional electronic and mechanical computers.[1] The simplest computational metamaterial systems have been used to implement fast, power efficient filtering operations in signal processing, as well as to integrate ordinary differential equations, among other applications.[1,5] These methods typically suffer from instability in their output due to fluctuations in the materials themselves, and can have rigid requirements for the materials in order to achieve practical levels of signal to noise.

In the context of machine learning, computational metamaterials can be advantageous in a pipeline involving analog sensors such as the MAPs listed above; ML can also improve the design of metamaterials, although this perspective is not explored in this paper. One subset of the metamaterials-based ML pipeline is the *encoding* of source information to be machine-learned. Carried by a wave field, the information is either contained in the spatial degrees of freedom of the wave or in its frequency degrees of freedom  Metamaterials contain dispersion in both domains and can thus process both spatial and temporal information. The second subset, *learned sensing*, invites the new engineering physics approaches alluded to earlier: that is, measurement and processing are jointly learned in order to only acquire and process the information that is needed for the given task, thus optimizing the entire sensing cycle.

We argue that the computing material and the input-output "data" be matched for the engineering problem at hand.[5] While we are not the first to make this argument, we do suggest here that advances in AI, simulation, and fabrication technologies,[21] along with theoretical and proof-of-principle advances,[8,9,22] make building physical computers a practically achievable target for widespread use in designing the next generation of  self-driving laboratories.

*Assertion 2: The existing general-purpose computing workflows for scientific laboratories and facilities are guaranteed suboptimal for AI-driven science and automation of high-throughput SDLs. Physical computation allows for bespoke and automatic design of optimal computational solutions for specific, rapid, and accurate measurement and processing of high-throughput analysis of materials (although at the cost of general computing capabilities).*

*Proposition 2:  We look to eschew prior practice for general digital computing and legacy laboratories, to design bespoke physical computers for precise, power efficient, parallelizable, real-time data processing pipelines in self-optimizing MAPs.*

Overall we are arguing for the development of a new kind of analog computing based on connectionist/distributed representations—a key part of what has given artificial neural networks their power—but exploiting the natural behavior of physical materials and mechanisms to do the computations. The several examples above introduce relaxation of a system following perturbation (*Physics-driven contrastive learning)*, wave propagation in a heterogeneous medium (*Wave-RNNs*), materials that learn to sense and encode data (*Metamaterials for ML*), and nonlinear transformations through mechanical, optical, and electrical systems (*Physics-aware training, PNNs*). The four are universal physical phenomena found in enumerable materials and systems, and are phenomena which are trainable using simulations.  That is, we can exploit recent advances in *simulation intelligence* (SI), a field that integrates simulation, artificial intelligence, and engineering physics,[23] carried out on general purpose computers to design bespoke physical computers. Physical computations can maintain the desirable properties of classic analog computers—essentially zero latency and near-zero information loss—but sidestep some of their main difficulties: classic electronic-analog circuits are highly sensitive to operating conditions, and require substantial human expertise to construct.[5] The four physical computing methods we highlight also share properties or mechanisms for uncertainty reasoning, leading to more reliable and robust MAPs. Recall that data flow in computational metamaterials amounts to a wave field that encodes for both spatial and temporal domains. This allows for certain parameterizations, for example the Green's function approach,[20] to quantify the already disentangled *aleatoric uncertainties*—stochasticities in data due to observation and process noise— which is nontrivial in digital computing and ML settings. And in an environment with PNNs, there could be many mechanical, optical, and electrical systems acting in concert but with mixed data types and noise, implying a normalized measure of uncertainty should be estimated and propagated to resolve the confidence of one physical system's output being fed into anothers.

Just like the methods we discuss here, distributed representations and the use of physics-for-calculations are central to biological systems in order to self-tune cellular machinery, execute metabolic processes, and even contribute to motivated behavior.[24-26] Unlike past efforts at analog circuits with biologically-inspired computation,[27] we do not use biological mediums (as in DNA computing) nor to mimic biological circuits in simulation (i.e., neuromorphic computing).[28] Instead, we recognize that the fundamental benefits, namely distributed representations and physics-for-calculations, are grounded mathematical ideas that can be realized directly in many diverse modes of solid-state materials.. Correspondingly for PC-MAP, computations can be programmed into a nearly limitless range of materials, and gradient-based learning can be executed on any sequence of physical input-output transformations, directly *in situ*.

## Examples

Key advantages to physical computers in the context of MAP include their implicit ability to minimize information loss in sensing, data processing, and optimization problems of physical components, speed at inference time, robustness in harsh environments (relative to the brittleness of integrated circuits), and compact form-factor (or "minaturizablity" relative to analog computing in general). Inference time, power efficiency, and limits on minaturizablity are also amongst the constraints facing practical deployment of many of today's emerging materials applications, including flow cell control for chemical, colloidal, and nano-partical synthesis (discussed in detail below) and in helping to predict online performance and failures online in self-healing in smart batteries.[29]

### *Inverse design of metamaterials*

An *inverse problem* is one of inferring hidden states or parameters from observations or measurements. If the forward simulation (or data-generation) process is $x \rightarrow y$, we seek to infer $x$ given $y$. In scientific settings, an inverse problem is the process of calculating from observations the causal factors that produced them. In practice this is non-trivial, as one needs to derive a forward model (i.e., a simulation or a mathematical/physical model) to predict data given an underlying objective, and then compute the mathematical inverse of the forward model via nonlinear programming and optimization with regularization. Recently, deep learning-based approaches provide a more promising way to inverse problem solving when conventional mathematical inversion is challenging. Rather than relying entirely on an accurate physics or mathematical model, the new data-driven machine learning methods can leverage large datasets to learn the inverse mapping end-to-end.[23]

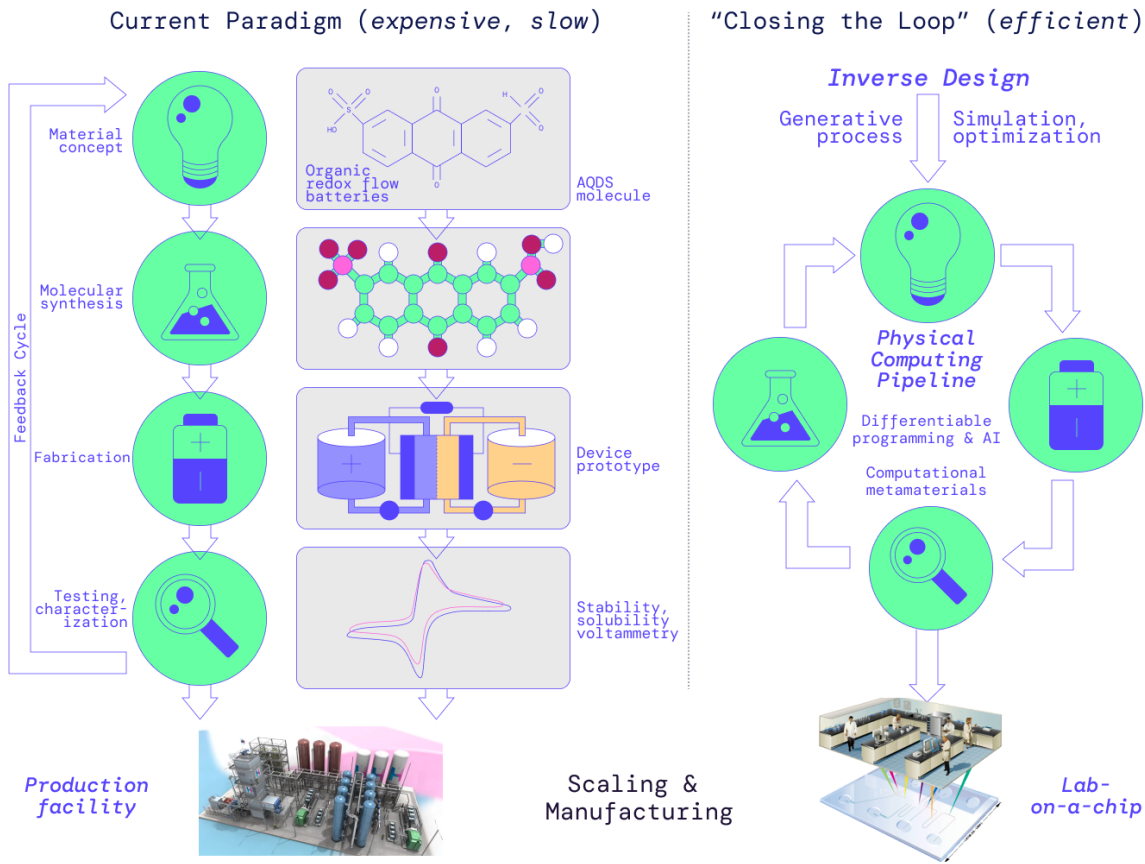

**Figure 2, Materials discovery as a closed-loop** — In a typical MAP with 3D flow cell chemistry,[30] the degree of miniaturization is limited by mechanisms for control (DAC) and analysis. Physical computers, however, may be printed directly into the flow cell making the whole "lab" tiny and therefore highly scalable (bottom right). (Left) The current material discovery paradigm (here for organic redox flow batteries) can be reworked to a closed-loop, inverse design counterpart *in silico* that maximizes efficiency and minimizes *in situ* resource-utilization (right) — to maximize this potential calls for physical learning and optimization in the closed-loop.[23]

Beyond the classic inverse problems, such as image reconstruction, inverse scattering, and computed tomography, much recent effort has been dedicated to the design of new materials and devices, and to the discovery of novel molecular, crystal, and protein structures via inverse processes. This can be described as a problem of *inverse design:* discovering the optimal structures or parameters of a design based on the desired functional characteristics — for example, computational inverse design approaches are finding many use-cases in optimizing nanophotonics[11,31,32] and crystal structures,[33–35] which can directly improve efficiencies of graphene and other 2D materials for photovoltaics and other renewable energy absorption/transfer surfaces.[36] Critical to our approach is the use of inverse design to design not new materials but new computing materials—for photonic computing,[37–39] electrical surface wave propagation, or audio processing.

Fig. 2 illustrates an example of how inverse design cycles can replace the existing, deductive paradigm (that can be expensive and inefficient) and yield suboptimal solutions. The diagram also elucidates how the physical setup to support inverse design processes in MAPs is quite different from existing labs and processes. With the introduction of physical computing, the optimization process can in itself be a fully miniaturized closed-loop optimization. Physical computing implemented as in Wright et al.[4] with a MAP inverse design workflow like Fig. 2, we get never-before-possible ways to optimize the MAP system. The most material and cost efficient route to new materials discovery is the use of 3d printing technique for microfluidics synthesis systems. Properly designed physical computing systems can be directly embedded in these miniaturized flow systems allowing for the synthesis, ML-based data analysis, and ML-directed candidate generation to happen on the same chip. The result is a truly scalable self-driving laboratory in terms of power use, physical space, material usage, data overhead, and inference time. Further, tight integration via miniaturization of synthesis, sensors (discussed below), and inference may even allow existing robotic self-driving laboratories, like AMADA,[40] to be miniaturized into lab-on-a-chip systems.

In vast, complex spaces like materials science, inverse design can yield de novo molecules to help in humanity's greatest challenges; according the DOE's 2019 AI for Science Report,[41] "AI amenable to inverse design will enable 'hypothesis discovery' and reduce our collective reliance on human intuition, with the potential to accelerate the pace of biological and environmental sciences by orders of magnitude." And when embedded with physical computing, we suggest MAPs can crank out another magnitude of efficiency in AI-driven science.

*ML-enabled intelligent sensor design*

Taking this concept of optimizing a system of cyber-physical components to a smaller scale of MAPs — i.e. more precisely at the level of measurement devices — we aim for *ML-enabled intelligent sensor design*: the use of inverse design and related machine learning techniques to optimally design data acquisition hardware jointly for both sensing, and analysis, via user-defined cost function or design constraints. This can directly resolve the existing MAP challenges of guaranteed suboptimal devices and cyber-physical information loss that we described earlier. Examples range from point-of-care diagnostics and on-chip spectroscopy, to biosensing with RNA "switches" and recognizing humans with triboelectric flooring.[42] Powered by physical computing we can monitor and tune existing processes (like material synthesis steps) and design new machinery, leading to very efficient implementations in terms of energy spent and inference time, as well as reliability and suitability to harsh environments. Please see Lavin et al.[23] for details on the process.

# Discussion

There is untapped potential for optimizing hardware in sync with material discovery that relies on said hardware. Such a positive feedback loop can advance the fields of AI, simulation, scientific computing, and engineering physics in tandem. Without which, the MAP field will settle into a guaranteed-suboptimal regime of general-purpose digital computing and scientific methods. Consider the opportunity for synchronous cyber-physical optimization, where, for example, physical neural networks can facilitate diverse computing environments with a variety of machine-types and sensor systems. It is not unreasonable to expect unconventional machine learning hardware as a result that is orders of magnitude faster and more energy efficient than conventional electronic processors. Synchronous cyber-physical optimizations like this can push beyond the simple optimization problems and human biases ingrained in today's scientific and engineering physics workflows, and to develop cyber-physical systems that can be used for scientific discovery — i.e., the MAPs that are capable of truly closed-loop operation, acquiring and generating knowledge.[21,23]

### Ethical considerations

While expecting significant advances with our proposal for MAPs and self-driving labs, it is essential to track additional ethical and societal implications that may arise. For instance, automation bias and algorithm aversion can appear in varied ways when exploring new dimensions of scientific methods with simulation and AI technologies. High costs associated with data collection, data access, and computing power are already problematic in today's society, particularly in creating and exacerbating inequalities in research output between state and private institutions as well as between developed and developing nations. The discovery of new materials in biological, chemical, and environmental realms can compound these worries; societal implications are pervasive in potential applications, such as biofuels and zero-carbon cement, novel energy production and storage such as nuclear fusion/fission and geothermal, and Earth systems deployments like carbon capture and modifying crops, all of which can have (unforeseen) downstream effects. And while AI and MAPs can lead to breakthrough medicines and energies, the same technologies and discoveries can be misused for de novo design of biochemical weapons—we must be aware of and have countermeasures against such dual-use activities.[43]

### Conclusion

*"We can not solve our problems with the same level of thinking that created them." -Einstein*

The role of SDL the next several years and several decades for biological, chemical, and environmental sciences is fundamental.[41] The role of SI in shaping new dimensions of the scientific method can have similar impact,[23] with an emphasis on physical computing, inverse design, and others that define new classes of engineering physics. The PC-MAP vision we've introduced opens up new possibilities for computational architectures based not on general abstractions but on concrete physics. We put forth a highly promising new class of engineering physics with immediate and long-term value for MAPs designs, en route to self-driving labs that enable new types of scientific methods.

# Acknowledgments & Resource Availability

*Lead contact*

Further information and requests for resources and materials should be directed to and will be fulfilled by the lead contact, Alexander Lavin (lavin@simulation.science).

*Materials availability*

This study did not generate new reagents.

*Data and code availability*

This study does not report original code, nor was data obtained, generated, or used for this (preliminary) study. Any additional information required to reproduce methods and results highlighted in the paper is available from the lead contact upon request.

**Author Contributions**

E.P. and A.L. shared in all parts of project conceptualization, methodology, investigation, analysis, and manuscript writing, reviewing, and editing.

**Declaration of interests**

The authors declare no competing interests.

# References

1. Zangeneh-Nejad, F., Sounas, D.L., Alù, A., and Fleury, R. (2021). Analogue computing with metamaterials. Nature Reviews Materials *6*, 207–225.

2. Flores-Leonar, M.M., Mejía-Mendoza, L.M., Aguilar-Granda, A., Sanchez-Lengeling, B., Tribukait, H., Amador-Bedolla, C., and Aspuru-Guzik, A. (2020). Materials Acceleration Platforms: On the way to autonomous experimentation. Current Opinion in Green and Sustainable Chemistry *25*, 100370.

3. Bhatt, U., Antorán, J., Zhang, Y., Vera Liao, Q., Sattigeri, P., Fogliato, R., Melançon, G., Krishnan, R., Stanley, J., Tickoo, O., et al. (2021). Uncertainty as a Form of Transparency: Measuring, Communicating, and Using Uncertainty. Proceedings of the 2021 AAAI/ACM Conference on AI, Ethics, and Society.

4. Wright, L.G., Onodera, T., Stein, M.M., Wang, T., Schachter, D.T., Hu, Z., and McMahon, P.L. (2022). Deep physical neural networks trained with backpropagation. Nature *601*, 549–555.

5. MacLennan, B.J. (2014). The promise of analog computation. International Journal of General Systems *43*, 682–696.

6. Doyne Farmer, J., Los Alamos National Laboratory, and Los Alamos National Lab. , NM (USA) (1989). A Rosetta Stone for Connectionism.

7. Movellan, J.R. (1991). Contrastive Hebbian Learning in the Continuous Hopfield Model. Connectionist Models, 10–17.

8. Dillavou, S., Stern, M., Liu, A.J., and Durian, D.J. (2022). Demonstration of Decentralized Physics-Driven Learning. Physical Review Applied *18*.

9. Hughes, T.W., Williamson, I.A.D., Minkov, M., and Fan, S. (2019). Wave physics as an analog recurrent neural network. Sci Adv *5*, eaay6946.

10. Hooker, S. (2021). The hardware lottery. Communications of the ACM *64*, 58–65.

11. Molesky, S., Lin, Z., Piggott, A.Y., Jin, W., Vucković, J., and Rodriguez, A.W. (2018). Inverse design in nanophotonics. Nature Photonics *12*, 659–670.

12. Peurifoy, J., Shen, Y., Jing, L., Yang, Y., Cano-Renteria, F., DeLacy, B.G., Joannopoulos, J.D., Tegmark, M., and Soljačić, M. (2018). Nanophotonic particle simulation and inverse design using artificial neural networks. Sci Adv *4*, eaar4206.

13. Stern, M., Arinze, C., Perez, L., Palmer, S.E., and Murugan, A. (2020). Supervised learning through physical changes in a mechanical system. Proceedings of the National Academy of Sciences *117*, 14843–14850.

14. Mouret, J.-B., and Chatzilygeroudis, K. (2017). 20 years of reality gap. Proceedings of the Genetic and Evolutionary Computation Conference Companion.

15. Howison, T., Hauser, S., Hughes, J., and Iida, F. (2020). Reality-Assisted Evolution of Soft Robots through Large-Scale Physical Experimentation: A Review. Artif. Life *26*, 484–506.

16. Degrave, J., Hermans, M., Dambre, J., and Wyffels, F. (2019). A Differentiable Physics Engine for Deep Learning in Robotics. Front. Neurorobot. *13*, 6.

17. Zhou, F., and Chai, Y. (2020). Near-sensor and in-sensor computing. Nature Electronics *3*, 664–671.

18. Martel, J.N.P., Muller, L.K., Carey, S.J., Dudek, P., and Wetzstein, G. (2020). Neural Sensors: Learning Pixel Exposures for HDR Imaging and Video Compressive Sensing With Programmable Sensors. IEEE Trans. Pattern Anal. Mach. Intell. *42*, 1642–1653.

19. Mennel, L., Symonowicz, J., Wachter, S., Polyushkin, D.K., Molina-Mendoza, A.J., and Mueller, T. (2020). Ultrafast machine vision with 2D material neural network image sensors. Nature *579*, 62–66.

20. Silva, A., Monticone, F., Castaldi, G., Galdi, V., Alù, A., and Engheta, N. (2014). Performing mathematical operations with metamaterials. Science *343*, 160–163.

21. Seifrid, M., Hattrick-Simpers, J., Aspuru-Guzik, A., Kalil, T., and Cranford, S. (2022). Reaching critical MASS: Crowdsourcing designs for the next generation of materials acceleration platforms. Matter *5*, 1972–1976.

22. Lin, M.M. (2020). Circuit Reduction of Heterogeneous Nonequilibrium Systems. Phys. Rev. Lett. *125*, 218101.

23. Lavin, A., Krakauer, D., Zenil, H., Paige, B., Baydin, A.G., Prunkl, C., Isayev, O., Macke, J., Cranmer, K., Hanuka, A., et al. (2021). Simulation Intelligence: Towards a New Generation of Scientific Methods. Preprint at arXiv, 10.48550/arXiv.2112.03235.

24. Brangwynne, C.P., Koenderink, G.H., MacKintosh, F.C., and Weitz, D.A. (2008). Cytoplasmic diffusion: molecular motors mix it up. J. Cell Biol. *183*, 583–587.

25. Nicholson, D.J. (2019). Is the cell really a machine? Journal of Theoretical Biology *477*, 108–126.

26. Battle, C., Broedersz, C.P., Fakhri, N., Geyer, V.F., Howard, J., Schmidt, C.F., and MacKintosh, F.C. (2016). Broken detailed balance at mesoscopic scales in active biological systems. Science *352*, 604–607.

27. Amant, R.S., St. Amant, R., Yazdanbakhsh, A., Park, J., Thwaites, B., Esmaeilzadeh, H., Hassibi, A., Ceze, L., and Burger, D. (2014). General-purpose code acceleration with limited-precision analog computation. 2014

ACM/IEEE 41st International Symposium on Computer Architecture (ISCA).

28. Grozinger, L., Amos, M., Gorochowski, T.E., Carbonell, P., Oyarzún, D.A., Stoof, R., Fellermann, H., Zuliani, P., Tas, H., and Goñi-Moreno, A. (2019). Pathways to cellular supremacy in biocomputing. Nat. Commun. *10*, 5250.

29. Vegge, T., Tarascon, J., and Edström, K. (2021). Toward Better and Smarter Batteries by Combining AI with Multisensory and Self-Healing Approaches. Advanced Energy Materials *11*, 2100362.

30. Sanchez-Lengeling, B., and Aspuru-Guzik, A. (2018). Inverse molecular design using machine learning: Generative models for matter engineering. Science *361*, 360–365.

31. So, S., Badloe, T., Noh, J., Bravo-Abad, J., and Rho, J. (2020). Deep learning enabled inverse design in nanophotonics. Nanophotonics *9*, 1041–1057.

32. Zhang, J.S.B. (2021). A directional Gaussian smoothing optimization method for computational inverse design in nanophotonics. Materials and Design *197*, 109213.

33. Noh, J., Kim, J., Stein, H.S., Sanchez-Lengeling, B., Gregoire, J.M., Aspuru-Guzik, A., and Jung, Y. (2019). Inverse Design of Solid-State Materials via a Continuous Representation. Matter *1*, 1370–1384.

34. Noh, J., Gu, G.H., Kim, S., and Jung, Y. (2020). Machine-enabled inverse design of inorganic solid materials: promises and challenges. Chem. Sci. *11*, 4871–4881.

35. Fung, V., Zhang, J., Hu, G., Ganesh, P., and Sumpter, B.G. (2021). Inverse design of two-dimensional materials with invertible neural networks. npj Computational Materials *7*.

36. Chen, Y., Zhu, J., Xie, Y., Feng, N., and Liu, Q.H. (2019). Smart inverse design of graphene-based photonic metamaterials by an adaptive artificial neural network. Nanoscale *11*, 9749–9755.

37. Lin, X., Rivenson, Y., Yardimci, N.T., Veli, M., Luo, Y., Jarrahi, M., and Ozcan, A. (2018). All-optical machine learning using diffractive deep neural networks. Science *361*, 1004–1008.

38. Zuo, S.-Y., Tian, Y., Wei, Q., Cheng, Y., and Liu, X.-J. (2018). Acoustic analog computing based on a reflective metasurface with decoupled modulation of phase and amplitude. Journal of Applied Physics *123*, 091704.

39. Pors, A., Nielsen, M.G., and Bozhevolnyi, S.I. (2015). Analog computing using reflective plasmonic metasurfaces. Nano Lett. *15*, 791–797.

40. Wagner, J., Berger, C.G., Du, X., Stubhan, T., Hauch, J.A., and Brabec, C.J. (2021). The evolution of Materials Acceleration Platforms: toward the laboratory of the future with AMANDA. Journal of Materials Science *56*, 16422–16446.

41. Stevens, R., Taylor, V., Nichols, J., Maccabe, A., Yelick, K., and Brown, D. (2020). AI for Science: Report on the Department of Energy (DOE) Town Halls on Artificial Intelligence (AI) for Science.

42. Ballard, Z., Brown, C., Madni, A.M., and Ozcan, A. (2021). Machine learning and computation-enabled intelligent sensor design. Nature Machine Intelligence *3*, 556–565.

43. Urbina, F., Lentzos, F., Invernizzi, C., and Ekins, S. (2022). Dual use of artificial-intelligence-powered drug discovery. Nature Machine Intelligence *4*, 189–191.

44. Humphreys, D., Kupresanin, A., Boyer, M.D., Canik, J., Chang, C.S., Cyr, E.C., Granetz, R., Hittinger, J., Kolemen, E., Lawrence, E., et al. (2020). Advancing Fusion with Machine Learning Research Needs Workshop Report. Journal of Fusion Energy *39*, 123–155.

# Appendix

*Risks*

In addition to ethical risks introduced above, we highlight several of the PC-MAP proposal's main technological and operational risks and mitigations in the table below:

**Table A1, Top risks and mitigations**

| Risk | Mitigation(s) | Value* |
|------|---------------|--------|
| ***Undersupply of MAP hardware to develop and test with*** — We anticipate significant development to be done in simulation platforms such as Unity, Nvidia Isaac, MoveIt, etc., but the transition to real hardware (let alone robust validation runs) may be challenging because robotics (and wetlabs) cost upwards of several million dollars, and labs may not risk a closed-loop self-optimization scheme with such expensive hardware. | - Co-development partnerships with robotics companies, including the newer cloud labs<br>- Collaborations with larger university labs<br>- Grants or CRADA with government | 0.3 x 7 = 2.1 |
| ***Limitations of bespoke PC-MAPs*** — A limit of physical computers is they are naturally bespoke; in designing them, we exchange general computing capabilities for high-speed computing in matter itself, without the implementation overhead and information loss implicit in traditional computing architectures when, that is, they simulate physical systems. | This trade-off is acceptable — implies one paradigm can't replace the other but that for select problems where speed,simplicity of validation and robustness are paramount, then material computing is a potentially transformative approach. | 0.5 x 5 = 2.5 |
| ***Expensive ramp-up drains resources before first deliverables or value inflection*** — many of the challenges to innovation in biotech and pharma industries are relevant here, even more so when considering the likely investment in hardware necessary ahead of generating valuable IP. | - Strategic mix of venture capital and corporate investment<br>- Align early wins/deliverables with revenue generating channels — get money inbound early | 0.4 x 9 = 3.6 |

\* Expected value is estimated as (probability of event) x (cost of event), which are [0,1] and [1,10], respectively. Thus high-risks (red) are [8.0,10.0], followed by mid (yellow) and low (green).

*Roadmap*

Towards realizing the outsized impact that physical computing can have on MAPs and more broadly self-driving labs, we briefly introduce several potential paths to convey a spectrum of estimated cost, time, and deliverables:

A. Minimal viable support: similar to a small-medium academic lab, this path is defined by small one-off contributions. Existing MAPs largely fall in this category.

B. Narrow, app-specific end-to-end build of a PC-MAP pipeline, open-sourced for others to mimic. AlphaFold, for example, would be a best-case scenario.

C. Grant-based (gov't or donors) platform development and deployment

D. Startup commercial platform

E. Incumbent — be it corporate or government, the incentives to share findings let alone open-source the MAP technologies are not there.

Our suggestion is proceeding with D or C; both can produce the requisite MAP-driving deliverables *and* a significant open-source MAP ecosystem, but with different pros/cons on capital efficiency and long-term R&D targets. Pasteur Labs is a *public-benefit* company for this reason.

**Success criteria** can be viewed as both near-term key-differentiators — *what MAP proficiency delta does PC provide?* — and the longer term upsides in use-inspired research — *what society-shaping technologies are made possible or better with PC-MAPs?* — and fundamental research — *have we discovered new means of computing, or new engineering physics?*

Please inquire for a detailed breakdown of the PC-MAP roadmap.